\documentclass[]{scrartcl}

\usepackage{amsmath,amssymb,amsfonts}%
\usepackage{amsthm}%
\usepackage{mathrsfs}%
\usepackage{epstopdf}
\usepackage{graphicx}
\usepackage{authblk}
\usepackage{textcomp}%
\usepackage{manyfoot}%
\usepackage{booktabs}%
\usepackage{subcaption}%
\usepackage{bbm}
\usepackage{hyperref}
\usepackage{multirow}
\usepackage{xcolor}
\usepackage{acro}

\DeclareAcronym{geneos}{
	short=GENEOs,
	long=Group Equivariant Non-Expansive Operators,
}
\DeclareAcronym{xai}{
	short=XAI,
	long=eXplainable Artificial Intelligence,
}
\DeclareAcronym{ai}{
	short=AI,
	long=Artificial Intelligence,
}
\DeclareAcronym{md}{
	short=MD,
	long=Molecular Dynamics,
}
\DeclareAcronym{rmsd}{
	short=RMSD,
	long= Root Mean Squared Deviation,
}
\DeclareAcronym{hb}{
	short=HB,
	long= Hydrogen Bond,
}

\usepackage[numbers, sort&compress]{natbib}
\bibpunct[, ]{[}{]}{,}{n}{,}{,}

\makeatletter%
\def\NAT@def@citea{\def\@citea{\NAT@separator}}
\makeatother%

\theoremstyle{plain}
\newtheorem{theorem}{Theorem}[section]

\newtheorem{corollary}[theorem]{Corollary}

\theoremstyle{definition}
\newtheorem{definition}[theorem]{Definition}

\theoremstyle{remark}

\providecommand{\keywords}[1]
{
	\small	
	\textbf{\textit{Keywords---}} #1
}

\date{}

\begin{document}
	
\title{GENEOnet: Statistical analysis supporting explainability and trustworthiness.}

\author[1]{Giovanni Bocchi\thanks{Corresponding author. Email: \texttt{giovanni.bocchi1@unimi.it}}}
\author[2]{Patrizio Frosini}
\author[1]{Alessandra Micheletti}
\author[3]{Alessandro Pedretti}
\author[4]{Carmen Gratteri}
\author[5]{Filippo Lunghini}
\author[5]{Andrea R. Beccari}
\author[5]{Carmine Talarico}

\affil[1]{Dept. of Environmental Science and Policy, University of Milan}
\affil[2]{Dept. of Computer Science, University of Pisa}
\affil[3]{Dept. of Pharmaceutical Sciences, University of Milan}
\affil[4]{LIGHT S.c.a.r.l.}
\affil[5]{Dompé Farmaceuici S.p.A.}

\maketitle

\begin{abstract}
Group Equivariant Non-Expansive Operators (GENEOs) have emerged as mathematical tools for constructing networks for Machine Learning and Artificial Intelligence. Recent findings suggest that such models can be inserted within the domain of eXplainable Artificial Intelligence (XAI) due to their inherent interpretability. In this study, we aim to verify this claim with respect to GENEOnet, a GENEO network developed for an application in computational biochemistry by employing various statistical analyses and experiments. 

Such experiments first allow us to perform a sensitivity analysis on GENEOnet’s parameters to test their significance. Subsequently, we show that GENEOnet exhibits a significantly higher proportion of equivariance compared to other methods. Lastly, we demonstrate that GENEOnet is on average robust to perturbations arising from molecular dynamics. These results collectively serve as proof of the explainability, trustworthiness, and robustness of GENEOnet and confirm the beneficial use of GENEOs in the context of Trustworthy Artificial Intelligence.
\end{abstract}

\keywords{
GENEOs; sensitivity analysis; equivariance; robustness; feature importance;}

\section{Introduction}

The confluence of medicinal chemistry and computational chemistry, coupled with advancements in drug design and discovery~\cite{torres2019, lionta2014}, presents numerous opportunities to investigate and evaluate the transparency and trustworthiness of Artificial Intelligence (AI) applications. In recent years, there has been a significant surge in AI solutions addressing computational biochemistry challenges~\cite{nag2022}. These have included the development of algorithms for generating three-dimensional protein structures,~\cite{marcu2022, yim2024, loeffler2024} predicting protein-protein interactions~\cite{tsuchiya2022}, and forecasting protein-ligand interactions~\cite{crampon2022, clyde2023, sauer2022}. However, the pace at which explainable models or validations of the trustworthiness of such AI systems have been developed has not kept up with these innovations. As a result, questions regarding the transparency of these systems have often gone unanswered, leading to skepticism within certain segments of the scientific community~\cite{ivanenkov2024, buttenschoen2024}.

One pressing challenge in this field is protein pocket prediction~\cite{laurie2006, palma2023}, which involves identifying locations within the three-dimensional structure of a protein where small molecules known as ligands, which usually have the role of drugs, are likely to bind. Notably, numerous machine learning and AI solutions have been proposed to tackle this problem~\cite{deeppocket2022,deepsite2017,p2rank2018,nazem2021,zhangtwo2020,Zhangone2020,marta2020,jiang2019}. However, as far as we are aware, none of these approaches has prioritized the transparency or trustworthiness of their underlying algorithms.

In particular, protein pocket detection is a problem characterized by a notable geometrical property: if a protein structure undergoes a rigid motion, its pockets should not change apart from their location and orientation. Although such rigid motions are unlikely to occur in real-world scenarios, due to the dynamic nature of proteins, this property would be highly desirable for any model predicting protein pockets, in order to trust its predictions. Moreover, the dynamical behavior of proteins can be simulated using molecular dynamics (MD), a powerful tool that allows researchers to model complex protein movements~\cite{cerutti2019, bernardi2015, meuwly2019}. This capability enables us to evaluate the robustness and coherence of the predictions of a pocket detector in a biologically relevant context. Specifically, MD simulations can be used to simulate a sequence of small, non-rigid perturbations to the initial protein structure, effectively allowing to test the resilience of pocket detection algorithms in response to structural changes.

Recently, we proposed GENEOnet~\cite{geneonet2022}, a network-based approach for protein pocket detection that leverages Group Equivariant Non-Expansive Operators (GENEO). A key design feature of GENEOnet is its inherent geometrical coherence, known as equivariance, which ensures that the predicted binding sites behave coherently under rigid motions. Furthermore, because of the mathematical properties of GENEOs, we were able to develop GENEOnet with the aim of having an explainable and robust method for protein pocket detection. In particular, the relatively small number of learnable parameters in the network allows us to assign meaningful interpretations to each parameter, effectively treating them as feature importance indicators~\cite{lime2016, shap2017, giudici2021, giannotti2023}, which are one of the most prominent research areas inside the field of XAI. 

In this study, following an introduction to GENEOs and GENEOnet, we will discuss how the parameters of GENEOnet can be easily interpreted from a statistical perspective, highlighting their potential to serve as global feature importance explanations. Furthermore, we will evaluate the reliability of GENEOnet outputs in response to rigid perturbations of the input data, marking a significant distinction from the other AI methods considered. Remarkably, this approach allows for a clear assessment of the robustness and stability of GENEOnet predictions under such conditions. Additionally, by exploiting molecular dynamics simulations, we will also investigate the consistency of GENEOnet outputs in response to non-rigid perturbations, further fostering trust in its predictions.

\section{Group Equivariant Non-Expansive Operators}

Recent advancements have proposed Group Equivariant Non-Expansive Operators (GENEOs) as powerful mathematical instruments for constructing network models within machine learning and artificial intelligence applications. The current state of GENEO theory is particularly active, focusing on developing methods to generate GENEOs in various contexts~\cite{consgeneo2022, newmet2018, finrep2023, genperm2023} and exploring their generalizations~\cite{ferrari2023}. For a comprehensive understanding of GENEOs and their properties, please refer to~\cite{towtop2019}.

At a high level, GENEOs should be perceived as data processing agents, defined for working with functional data. The two key characteristics of such agents, equivariance and non-expansivity, are inherent in their name.

Equivariance assumes growing importance when handling data subject to geometric transformations~\cite{convnet2016, harmnet2017, worral2017, cohen2016}. Equivariance is defined in relation to a group of transformations of the data domain, known as an equivariance group; to be considered equivariant, an operator must commute with all transformations within the equivariance group. The most basic form of equivariance, named invariance, implies that the agent's outputs stay the same when inputs are obtained through a transformation belonging to the equivariance group. For example, if we have a point cloud derived from an object scan, the represented object remains unchanged regardless of the spatial orientation of the point cloud. This necessitates that an agent processing such point clouds should be invariant with respect to the group of rigid motions of the Euclidean space to produce reliable results. Ultimately, equivariance serves as a mean to incorporate relevant data symmetries within the processing pipeline for problems of interest.

Non-expansivity, additionally, ensures that GENEOs are 1-Lipschitz operators with respect to appropriate distances on the functional input and output space. Non-expansivity carries significant implications when studying the topological properties of the GENEO space~\cite[see][]{towtop2019}, but it can also be seen as a form of resistance to perturbations in the data. Stability is a rare property found in black box models, which have often been demonstrated to be unstable in ways that can be exploited to return inconsistent and unreliable results, such as in the case of adversarial attacks~\cite{advmed2019,advmed2023,advspeech2020,strike2019}.

In combination, equivariance and non-expansivity can provide a high level of trustworthiness to models built using GENEOs ~\cite{geneonet2022, scenenet2023, geneograph2024}: the variability derived from geometric transformations is controlled alongside robustness to minor perturbations in the inputs.

For sake of clarity, we will now shortly formalize the concept of group-equivariant non-expansive operator. The reader can refer to~\cite{towtop2019} for a comprehensive understanding of GENEOs and their properties.

We assume that a space $\Phi$ of functions from a set $X$ to $\mathbb{R}$ is given, together with a group $G$ of permutations of $X$, such that 
if $\varphi\in\Phi$ and $g\in G$ then $\varphi\circ g\in \Phi$.  We call the couple $(\Phi,G)$  \emph{perception pair}.  We also assume that $\Phi$ is endowed with the topology induced by the
distance defined by the $L_\infty$-norm $D_\Phi(\varphi_1,\varphi_2)=||\varphi_1 - \varphi_2||_{\infty}, \ \varphi_1,\varphi_2\in\Phi$.
Let us assume that another perception pair $(\Psi,H)$ is given,  with $\Psi$ endowed with the topology induced by the analogous $L_\infty$-norm distance $D_\Psi$, and let's fix a homomorphism
$T:G\to H$. 
\begin{definition}
A map $F:\Phi\to\Psi$ is called a \emph{group equivariant non-expansive operator (GENEO)} if the following conditions hold:
\begin{enumerate}
  \item $F(\varphi \circ g)=F(\varphi)\circ T(g)$ for every $\varphi\in\Phi$, $g\in G$ (equivariance);
  \item $\|F(\varphi)-F(\varphi')\|_\infty \le \|\varphi-\varphi'\|_\infty$ for every $\varphi,\varphi'\in\Phi$ (non-expansivity).
\end{enumerate}
\end{definition}

If we denote by $F_{all}$ the space of all GENEOs between $(\Phi, G)$ and $(\Psi, H)$ and we introduce the metric
$$
D_{GENEO}(F_1,F_2)=\sup_{\varphi\in\Phi}||F_1(\varphi)-F_2(\varphi)||_{\infty},\qquad \forall F_1,F_2\in F_{all}
$$
the following main properties of spaces of GENEOs can be proven (see \cite{towtop2019} for the proofs).
\begin{theorem} \label{teo-compact}
If $\Phi$ and $\Psi$ are compact,  then $F_{all}$ is compact with respect to the topology induced by $D_{GENEO}$.
\end{theorem}

\begin{corollary}\label{coroll-compact}
If $\Phi$ and $\Psi$ are compact with respect to the $\infty$-norms $D_\Phi$ and $D_\Psi$, respectively,  then $F_{all}$ can be $\varepsilon$-approximated by a finite set for any $\varepsilon > 0$.
\end{corollary}

\begin{theorem}\label{teo-convex}
If $\Psi$ is convex, then $F_{all}$ is convex.
\end{theorem}

Theorem \ref{teo-compact} and Corollary \ref{coroll-compact} demonstrate that when data spaces are compact, the space of GENEOs inherits this compactness, thus allowing for adequate approximation by a limited number of representatives, which simplifies the problem. According to Theorem \ref{teo-convex}, if the data space is convex, any convex combination of GENEOs will also be a GENEO. Consequently, when both compactness and convexity are present, one can easily construct any element of $F_{all}$ using only a finite set of operators. Moreover, the convex nature of $F_{all}$ assures a unique minimum for each strictly convex cost function within the GENEO space, facilitating the identification of an 'optimal GENEO'. These properties, when utilized wisely, enable the development of highly efficient neural networks characterized by reduced complexity—fewer nodes and layers—leading to models with enhanced interpretability due to their clear architecture. GENEOnet serves as an early example of constructing a neural network with the application of GENEOs.

\section{GENEOnet}

\begin{figure}[ht]
		\centering
		\begin{subfigure}{0.45\textwidth}
			\centering
			\includegraphics[width=\textwidth]{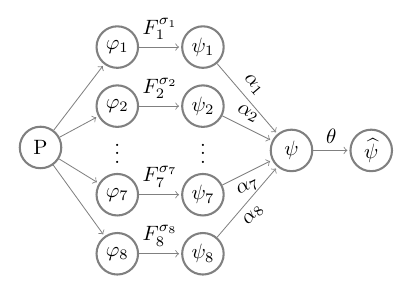}
			\caption{GENEOnet architecture.}
			\label{fig:model}
		\end{subfigure}
		\footnotesize
		\begin{subtable}{0.45\textwidth}
		\centering
			\begin{tabular}{ccccc}
				\toprule
				Unit & Channel & $\sigma$ & $\alpha$ & $\theta$\\
				\midrule
				1 &	Distance      &	3.110 &	0.362 &	0.756\\
				2 &	Gravitational &	5.197 &	0.002 &	\\
				3 &	Electrostatic &	2.561 &	0.054 &	\\
				4 &	Lipophilic	  & 4.678 &	0.338 &	\\
				5 &	Hydrophilic	  & 3.545 &	0.001 &	\\
				6 &	Polar	      & 6.166 &	0.185 &	\\
				7 &	HB Acceptor	  & 4.186 &	0.056 &	\\
				8 &	HB Donor	  & 3.908 &	0.001 &	\\
				\bottomrule
			\end{tabular}
			\caption{\label{tab:parameters} Optimal parameters' values.}
		\end{subtable}
		\caption{Model architecture and optimal parameters obtained after training the model.}	\label{fig:modeltab}
\end{figure}

GENEOnet~\cite{geneonet2022, geneonet2024} (which can be freely tested as a webservice at \href{https://geneonet.exscalate.eu}{https://geneonet.exscalate.eu}) is a specialized GENEO network model designed for protein pocket detection. It features a shallow network architecture that is composed of a limited number of GENEO units, as illustrated in Figure~\ref{fig:model}.

A brief summary follows of the six steps executed by GENEOnet to detect protein pockets (for further details refer to~\cite{geneonet2022}):
\begin{enumerate}
    \item Data preprocessing: On a protein $P$, GENEOnet first computes a grid discretizing the bounding box surrounding the protein surface. Subsequently, it approximates eight potential functions, each of which models essential aspects of the protein structure from geometric, physical, and chemical viewpoints. 
    \item GENEO layer: A convolutional rigid motion equivariant operator, each based on a kernel depending on a shape parameter $\sigma$, is applied to each potential function. The resulting function is normalized between 0 and 1.
    \item Convex combination: GENEO outputs are combined through a convex combination with weights $\alpha$. The resulting combination output, denoted as $\psi$, normalised between 0 and 1, represents the likelihood that each voxel belongs to a pocket.
    \item Thresholding: The final output is obtained by taking the connected components of the spatial region in which $\psi$ is above the parameter $\theta$.
    \item Evaluation: Having the ground truth (i.e. the ligand) available, the output can be compared using a volumetric accuracy function.
    \item Scoring: Each predicted pocket is assigned a score, computed as a weighted average of $\psi$ within its spatial region, enabling pocket ranking.
\end{enumerate}

Despite its relative simplicity and the quite small number of learnable parameters (the unknown parameters are only 17, as can be seen in Table \ref{tab:parameters}), GENEOnet has been shown~\cite[see][]{geneonet2022} to exhibit slightly superior performance in identifying the true pockets with the top-ranked predicted pockets when compared to state-of-the-art methods. 

One of the key advantages of GENEOnet lies in its minimal number of trainable parameters, a factor that enables the assignment of meaningful interpretations for humans to each of its seventeen parameters. The shape parameters $\sigma$ exert an influence over the operator kernels, while the convex combination coefficients $\alpha_i$ serve as feature importances for potentials (Table~\ref{fig:modeltab}). Lastly, the threshold coefficient $\theta$ determines the significance level for voxel activation.

The construction of the model with a reduced number of parameters has to be attributed to two primary factors: the equivariance property and the knowledge injection employed during the selection of potentials and corresponding operators. The employment of GENEOs that are equivariant with respect to rigid motions of the space significantly reduces the number of parameters because the model does not need to learn this symmetry from the data; instead, it is directly encoded in the choice of operators.

Furthermore, the convex constraint on the $\alpha$ coefficients may promote their sparsity during learning, thereby further minimizing the number of significant parameters. The convex combination parameters merit additional attention because, as the intermediate outputs $\psi_j$ are normalized between 0 and 1, the convex combination coefficients assume the role of feature importances for the various potentials and units within the GENEO layer. These feature importances constitute an  {\textit explanation} for this model yet independent of any specific instance of protein to which it is applied; thus, the $\alpha$ coefficients can be regarded as a {\textit model-specific global explanation} for GENEOnet (for a complete taxonomy refer to~\cite{guidotti2021}).

\section{Data sources}

Two publicly accessible datasets have been utilized for subsequent analyses: PDBbind~\cite{pdbbind2015} and ATLAS~\cite{atlas2023}. The PDBbind database consists of a curated collection of biomolecular complexes accompanied by experimentally determined binding affinities. The 2020 version contains approximately 19,000 protein-ligand complexes; nearly 12,000 have been chosen for the training and evaluation of GENEOnet. This same subset will be utilized also in the analyses performed in this work. Conversely, ATLAS contains MD simulations for 1,390 proteins, along with analyses of the dynamics and data visualizations. From ATLAS, we obtained 37 simulations corresponding to proteins that are shared with the aforementioned set extracted from the PDBbind and will be employed in the following sections.

\section{Sensitivity analysis}
\label{subsec:sensitivity}

A first statistical analysis was conducted to examine the response of GENEOnet parameters to variations in the training dataset.

\begin{figure}[ht]
        \centering
		\begin{subfigure}{0.45\textwidth}
			\centering
			\includegraphics[width=\textwidth]{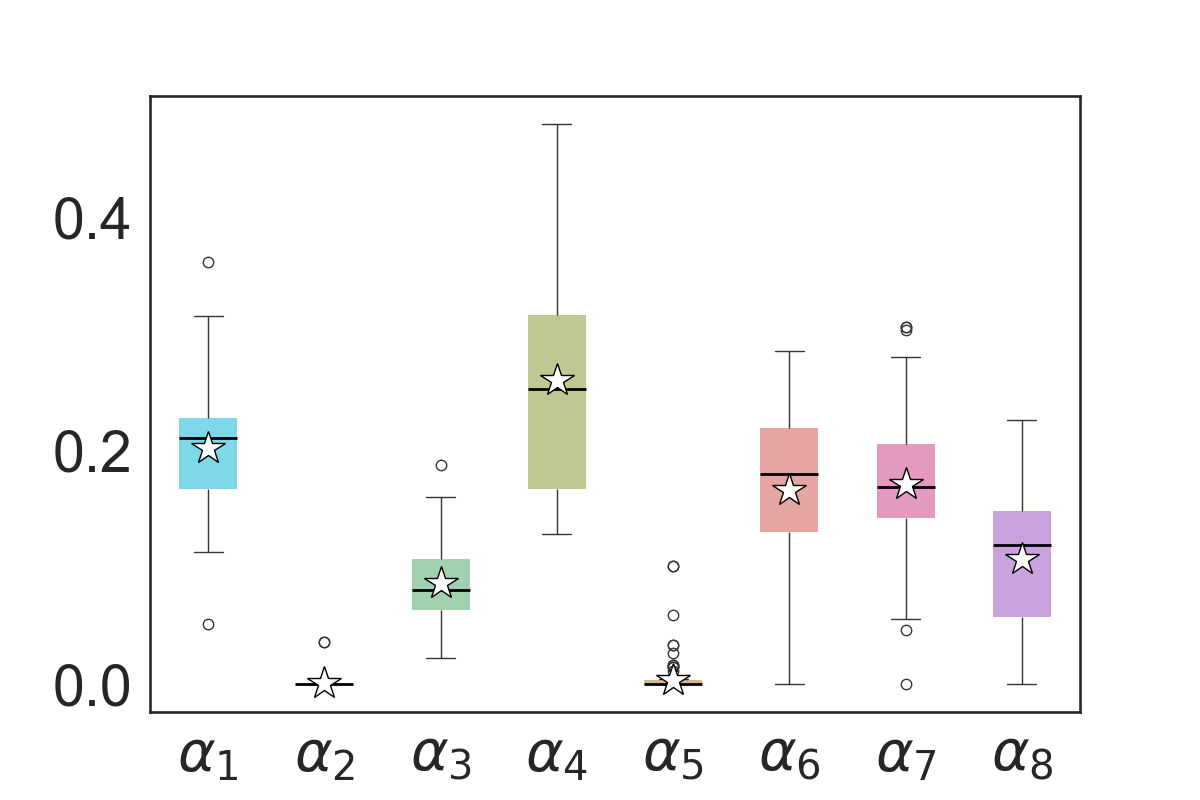}
			\caption{Distributions of convex combination coefficients.}
			\label{fig:sensitivity1}
		\end{subfigure}
		\begin{subfigure}{0.45\textwidth}
			\centering
			\includegraphics[width=\textwidth]{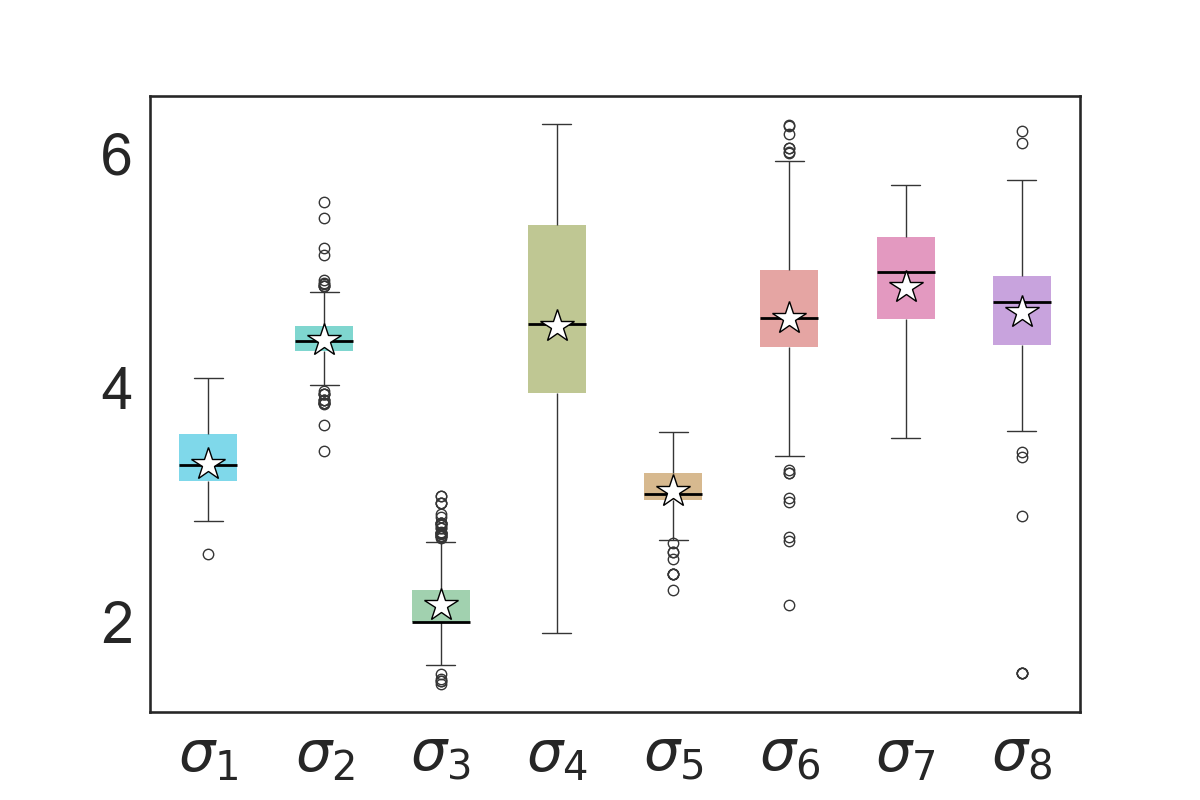}
			\caption{Distributions of convolutional kernel coefficients.}
			\label{fig:sensitivity2}
		\end{subfigure}
        \caption{Empirical distributions of the estimated parameters of GENEOnet, obtained by randomly varying the training set}
\end{figure}

The training set for GENEOnet consisted of 200 protein-ligand complexes; accordingly, this analysis involved $n = 200$ independent repetitions of the model training, with each iteration featuring a training set of 200 complexes selected uniformly at random from the data collected from the PDBbind. The training optimization always started from the same initial parameter guess in each repetition, ensuring that the only source of variation between the different runs was the training set itself. The boxplots displayed in Figure~\ref{fig:sensitivity1} and~\ref{fig:sensitivity2} thus represent sensitivity analyses of the parameters with respect to variations in the training data. 
  
The results reported in Figure \ref{fig:sensitivity1} show that $\alpha_1$ and $\alpha_4$, which are also the largest coefficients of the optimal set of convex combination parameters reported in Table \ref{fig:modeltab}, have distributions which are quite far from zero, thus showing that the corresponding potentials, the Distance and the Lipophilic one, maintain a high importance in the pocket identification procedure, whatever the training set of the model. In contrast, the distributions of $\alpha_2$ and $\alpha_5$ are concentrated on values close to zero, so the corresponding potentials, the Gravitational and Hydrophilic ones,  never play a relevant role in identifying pockets. 
This further strengthens the explanatory role of the parameters $\alpha$ compared to the mere point estimates shown in  Table~\ref{tab:parameters}. 

Furthermore, a biochemical rationale underpins some of these findings, as evidenced by the frequent negative correlation between Hydrophilic and Lipophilic potentials. Typically, the exposed regions of a protein surface are hydrophilic, given their solvation primarily in aqueous environments. Conversely, less exposed and more sheltered areas may exhibit hydrophobic (and thus lipophilic) characteristics. Consequently, the model incorporates analogous information through both the Lipophilic and Hydrophilic potentials. Thus, it is appropriate that only one of these is deemed significant for the identification of pockets that are usually unlikely to be placed in the most exposed regions of the protein surface.

The distributions of the coefficients of the convolutional kernels in Figure \ref{fig:sensitivity2} have a dispersion that is similar to the corresponding dispersion of the distributions of the convex combination coefficients, thus confirming the same uncertainty quantification of the point estimates of the parameters of each potential.

\subsection{Equivariance analysis}
\label{subsec:equivariance}

Equivariance is pivotal in understanding the functioning of GENEOnet, as it makes its predictions consistent with respect to rigid motions in $\mathbb{R}^3$ of the protein structure. It is fundamental in our example since the specific position in which the protein is observed should not change the pockets identification.  However, upon discussing the matter with practitioners of medicinal chemistry, we found that equivariance is not a feature that an average pocket identification algorithm will guarantee. Consequently, we decided to evaluate the level of equivariance displayed by GENEOnet and compare it with several state-of-the-art algorithms considered in~\cite{geneonet2022}.

To this aim, the methodology applied is the following:
\begin{enumerate}
    \item We systematically sampled $N = 2000$ proteins $(P_1, \dots, P_N)$ from the larger dataset extracted from PDBbind.
    \item For each protein $P_i$ we generated a rotated protein $\rho(P_i)$ obtained by applying to the structure a rotation $\rho$ of $\pi/2$ around the $x$-axis.
    \item For each method $\mathcal{M}$ under investigation, we computed predictions on both the original input $\mathcal{M}(P_i)$ and the rotated $\mathcal{M}(\rho(P_i))$.
    \item The analysis was confined to the three top-ranked pockets. We will denote as 
    $$\mathcal{M}(P_i)_j\quad j \in \{1,2,3\}$$
    the spatial region (reported on the GENEOnet grid so that there is a common reference for all methods) identifying the $j$-th predicted pocket by method $\mathcal{M}$.
    \item By applying the inverse rotation $\rho^{-1}$ to the predicted pockets returned on the rotated proteins, we can compute the proportion of overlap between the non-rotated and rotated predictions for protein $P_i$ (here $|\cdot|$ stands for the volume).
    $$
        O^{\mathcal{M}}_{j}(P_i) = \frac{|\mathcal{M}(P_i)_j \cap \rho^{-1}(\mathcal{M}(\rho(P_i))_j)|}{|\mathcal{M}(P_i)_j|}\quad j \in \{1,2,3\}
    $$
    If method $\mathcal{M}$ didn't output a prediction for one index $j$ on either the original or the rotated protein then the value of $O^{\mathcal{M}}_{j}(P_i)$ has been considered a missing value.
    \item Finally for each $j \in \{1,2,3\}$ we estimated the proportion $p_j^{\mathcal{M}; \tau}$ of proteins for which the overlap is non-missing and above a threshold $\tau$ as follows:
    $$
        \hat{p}_j^{\mathcal{M}; \tau} = \frac{1}{|I^{\mathcal{M}}_{j}|} \sum_{i \in I^{\mathcal{M}}_{j}} \mathbbm{1}\Bigl(O^{\mathcal{M}}_{j}(P_i) \ge \tau\Bigr),
    $$
     where we denoted by $I^{\mathcal{M}}_{j}$ the set of indices for which $O^{\mathcal{M}}_{j}(P_i)$ is non-missing.
     Empirical confidence intervals based on the sample have also been computed.
\end{enumerate}

\begin{figure}[ht]
		\centering
		\begin{subfigure}{0.48\textwidth}
			\centering
			\includegraphics[width=\textwidth]{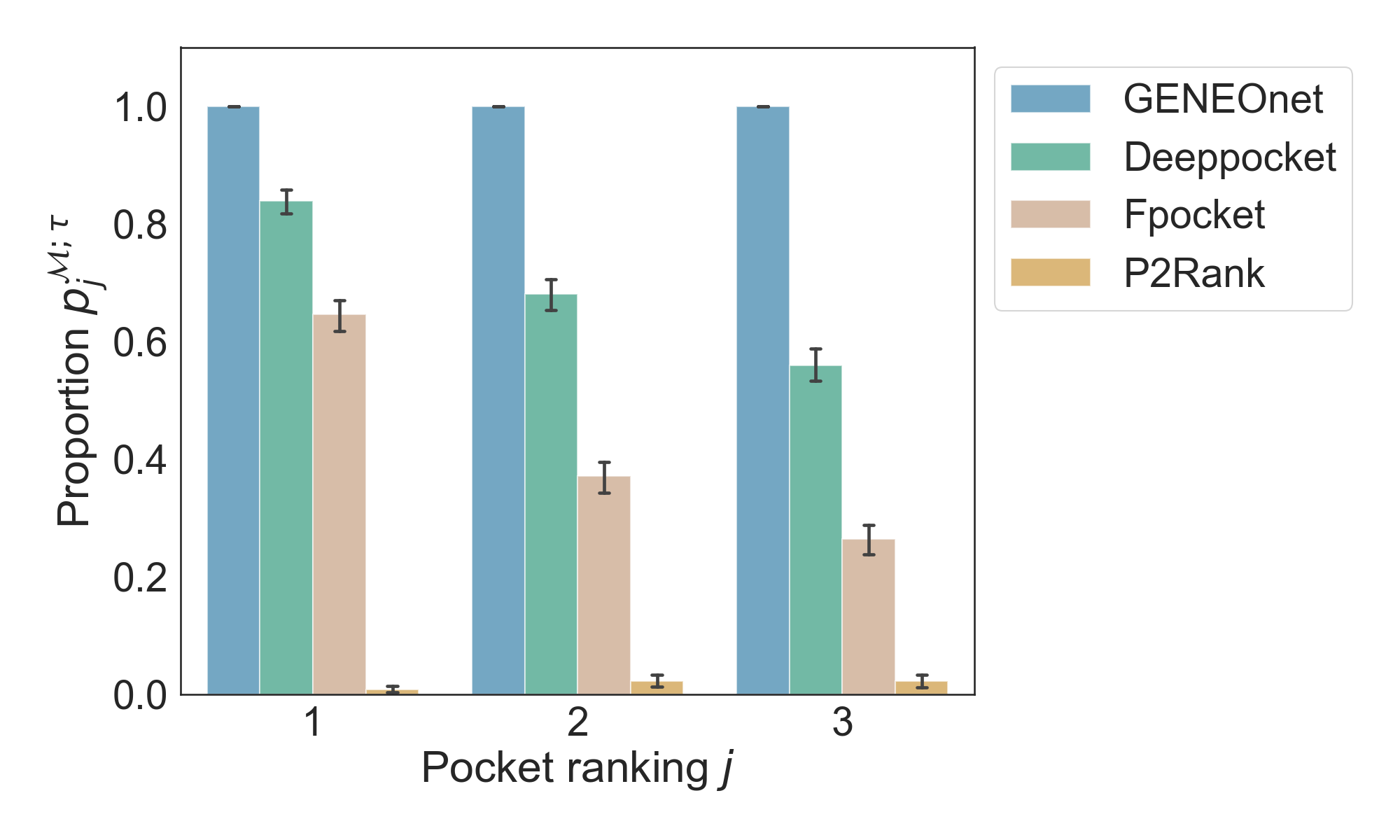}
			\caption{$\tau = 0.95$}
			\label{fig:equi1}
		\end{subfigure}
		\begin{subfigure}{0.48\textwidth}
			\centering
			\includegraphics[width=\textwidth]{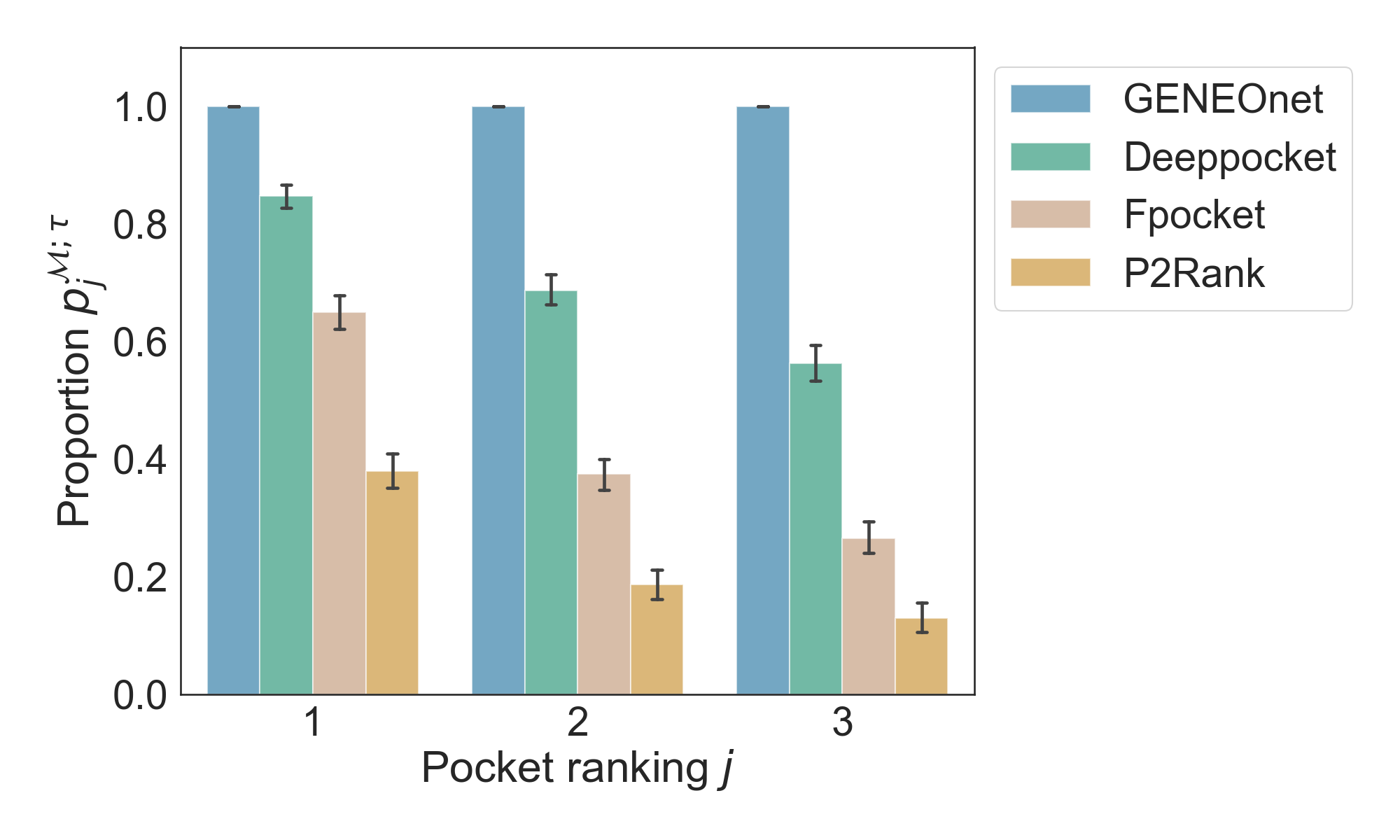}
			\caption{$\tau = 0.75$}
			\label{fig:equi2}
		\end{subfigure}\\
  		\begin{subfigure}{0.48\textwidth}
			\centering
			\includegraphics[width=\textwidth]{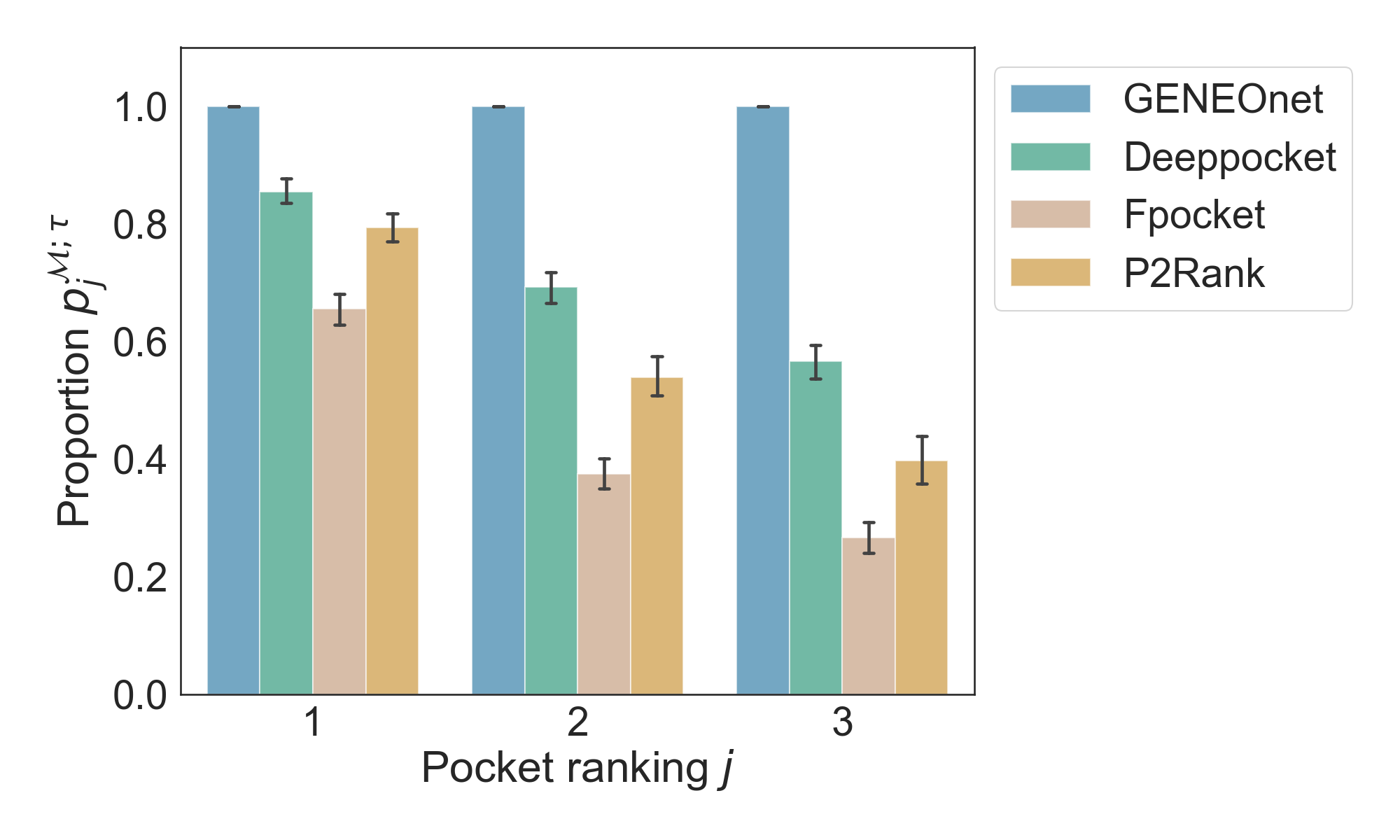}
			\caption{$\tau = 0.50$}
			\label{fig:equi3}
		\end{subfigure}
        \caption{Mean values and confidence intervals, obtained from different protein samples, of the proportions $\hat{p}_j^{\mathcal{M};\tau}$ computed for GENEOnet and other three methods for pocket detection, obtained for different values of the threshold $\tau$.}
\end{figure}

Setting $\tau=1$ allows for the estimation of the proportion of detected pockets that show perfect equivariance, while considering lower values of $\tau$ allows for differences that may arise from the implementation or numerical approximation errors, resulting in the calculation of the proportion of pockets showing approximate equivariance.
Figures~\ref{fig:equi1},~\ref{fig:equi2} and~\ref{fig:equi3} report estimates of $(p_1^{\mathcal{M}; \tau}, p_2^{\mathcal{M}; \tau}, p_3^{\mathcal{M}; \tau})$ for values of $\tau \in \{0.95, 0.75, 0.50\}$ and for the four methods: GENEOnet, DeepPocket~\cite{deeppocket2022}, Fpocket~\cite{fpocket2009} and P2Rank~\cite{p2rank2018} along with 99\% confidence intervals.

The results allow us to conclude that GENEOnet exhibits a higher degree of equivariance (approximately equal to 1) compared to other methods under consideration. The 99\% confidence intervals displayed in Figures~\ref{fig:equi1}, ~\ref{fig:equi2} and~\ref{fig:equi3} (and reported in Table~\ref{tab:equi}) demonstrate that the proportion of equivariance for GENEOnet is significantly greater than all other methods at a significance level of 0.01 for all the considered values of $\tau$, thus ranging from perfect to approximate equivariance. This is a significant finding as P2Rank and DeepPocket were, respectively, the second and third-best performing models in the task of pocket identification~\cite{geneonet2022}. However, their predictions are susceptible to some degree of instability with respect to the spatial pose of the input protein, which may also affect the pocket rankings when transitioning from the original to the rotated protein. This finding, which aligns with the claim made by medicinal chemistry experts, constitutes a significant source of unreliability for these models as the frozen pose, which is retrieved from databases such as the PDBbind and may be used in the training data, is far from being canonical in any sense. 

\subsection{Robustness analysis}
\label{subsec:robustness}

In the previous Section, we delved into testing the equivariance on protein structures subjected to rigid perturbations. However, it is crucial to extend our considerations also to non-rigid transformations, given the inherent dynamism of proteins that should be viewed as moving entities rather than static objects. In the realm of medicinal chemistry, MD simulations offer a valuable tool for studying the dynamic behavior of a protein immersed in a solvent environment, typically water molecules. 

Given that the time interval between successive frames in MD simulations is generally small, we can presume that structural changes between consecutive frames are minimal, apart possibly for a negligible number of cases.

With this rationale and owing to the intrinsic biological significance, we opted to assess the resilience of GENEOnet to detect the same pockets on protein MD data. 

For each of the 37 proteins sourced from the ATLAS database, we exploited two simulations spanning 100 ns each: the first composed of 1000 frames and the second consisting of 10000 (hence with a time step of 100 ps and 10 ps respectively). 

Upon collecting the data, we proceeded to execute the following experiments: 
\begin{enumerate}
    \item For each protein $P^i$ (where $i$ ranges from $1$ to $37$), we focused on the initial $T = 250$ frames of the dynamics, denoted as $P^i_t$, where $t$ spans from $1$ to $T$ (corresponding to 25 ns and 2.5 ns for each simulation, respectively)
    \item For each time step $t$, we computed the global prediction generated by GENEOnet (i.e. the union of all the pockets detected by GENEOnet on the surface of the protein), denoted as $\mathcal{G}(P^i_t)$, without segmenting it into distinct pockets, unlike the approach employed in the preceding Section.
    \item For every time step $t$ (where $t$ ranges from $2$ to $T$), we calculated the degree of overlap between the prediction generated for the current frame and that produced for the preceding frame:
    $$
        O_{t}(P^i) = \frac{|\mathcal{G}(P^i_{t-1}) \cap \mathcal{G}(P^i_{t})|}{|\mathcal{G}(P^i_{t-1})|}
    $$
    We also computed the root mean squared deviation (RMSD) between the atomic positions of the two consecutive frames considering all atoms:
    $$
        RMSD_{t}(P^i) = \sqrt{\frac{1}{|N_{i}|} \sum_{j = 1}^{N_{i}} \Bigl|\Bigl|x^j_{t-1} - x^j_{t}\Bigr|\Bigr|^2}
    $$
    where $N_{i}$ refers to the total count of atoms within protein $P^{i}$, while $x^j_t$ denotes the coordinate vector representing the position of atom $j$ at frame $t$.
\end{enumerate}

Assuming non-expansivity, we would expect to witness, to some degree, a negative correlation between the series representing RMSD and those depicting the overlap for the same protein. Actually, in some cases, a protein could move during the MD mainly in the parts where no pockets are present, thus it may happen to observe high RMSD values which do not correspond to small overlap values. The reverse should instead be observed more frequently, i.e. small RMSD values should correspond to high overlaps of the pockets.

\begin{figure}[p]
		\centering
		\begin{subfigure}{\textwidth}
			\centering
			\includegraphics[width=\textwidth]{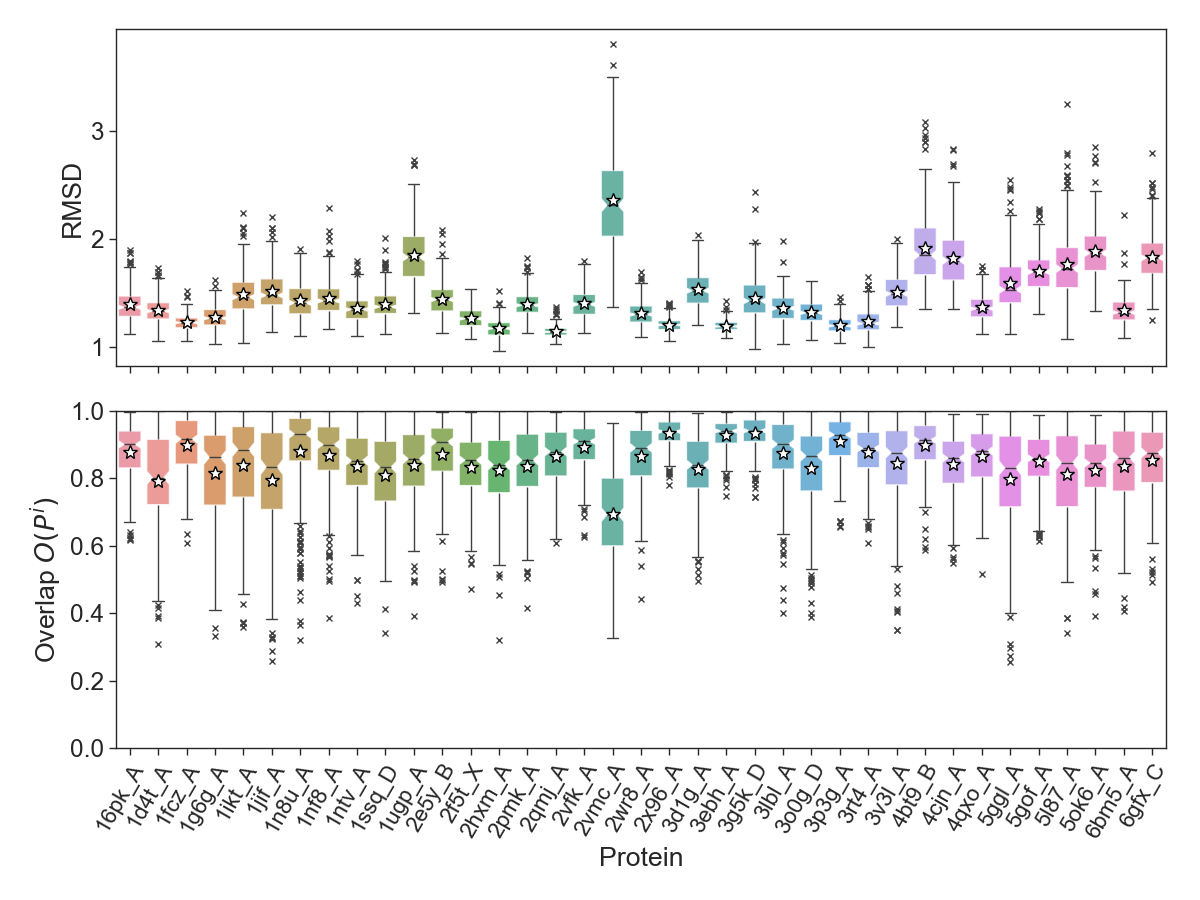}
			\caption{Distributions of overlap and RMSD for the dynamics with time delta 100 ps.}
			\label{fig:shorter}
		\end{subfigure}\\
		\begin{subfigure}{\textwidth}
			\centering
            \includegraphics[width=\textwidth]{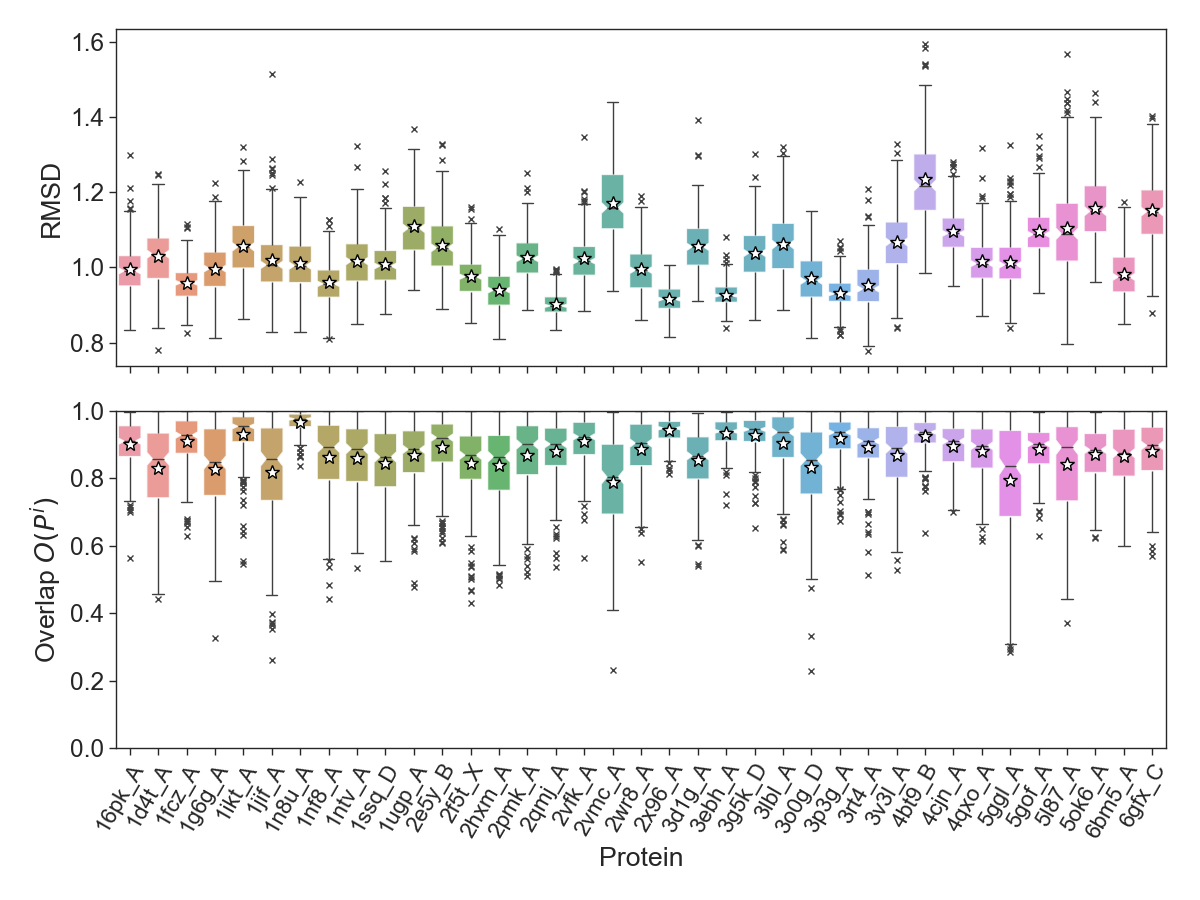}
			\caption{Distributions of overlap and RMSD for the dynamics with time delta 10 ps.}
			\label{fig:longer}
		\end{subfigure}
        \caption{Robustness analysis}
\end{figure}

Figures~\ref{fig:shorter} and~\ref{fig:longer} offer a comparative analysis of the distributions of overlaps and RMSD values, respectively, for the selected proteins for both molecular dynamics.

The negative correlation between overlap and RMSD is moderately clear in the simulations with a time delta of 100 ps (in Figure~\ref{fig:shorter}) since RMSD values are rarely below 1\r{A}. However, in some cases, higher RMSD distribution values correspond to small overlap distribution values, such as in the case of protein 2vmc\_A; but this is not always the case, as can be observed for proteins 1ugp\_A or 4bt9\_B. Reversely, when the distribution of RMSD is concentrated around small values, the corresponding distribution of the overlap is mainly located in correspondence of values close to 1, like, for example, for proteins 2x96\_A or 3ebh\_A.
 
The results are then in accordance with the definition of non-expansivity, which prescribes that the distance between the model outputs should be smaller than the distance between the inputs; thus pockets which are similar and close in two subsequent frames should be both detected and have a big overlap.

Moreover, in the case of finer MD simulations with a time delta of 10 ps we can make similar observations with the difference that the mean overlaps are generally higher and the mean RMSD tends to be lower as shown in Figure~\ref{fig:longer} (see also Appendix~\ref{app:perturbation} and Table~\ref{tab:meanoverlap} for a quantitative evaluation of overlaps).

\section{Concluding remarks}

In this study, we conducted a comprehensive evaluation of GENEOnet, a network architecture comprising GENEOs designed for protein pocket detection. To assess its explainability and trustworthiness, we employed several statistical analyses. 

Initially, a sensitivity analysis was conducted to assess the effect of variations in the training dataset on the model coefficients. This examination provided an enhanced comprehension of the parameters involved, facilitating improved interpretability and allowing for an evaluation of their statistical significance. 

Secondly, we compared the equivariance properties of GENEOnet with those of other state-of-the-art models for protein pocket detection. The results of this experiment demonstrated that GENEOnet's employment of GENEOs as building blocks yields a substantially higher proportion of equivariance than other methods, indicating that its outputs are significantly more reliable when subjected to rigid transformations of the input data. 

As a possible limitation of this work, we acknowledge that we considered only rotations $\rho$ of angle $\pi/2$ around the $x$ axis. Even if this was enough to obtain the results that were shown, it was done primarily in order to easily return the rotated voxel grid to the original grid for comparison. More general rotations (those with axes different from one of the coordinate ones, for example), instead, would have been harder to take into account because of the more challenging alignment of the grids that would be required in that case. In future work, we plan to test also other kinds of rotation or angles to further confirm the theoretical reliability of GENEOnet to these transformations.

Lastly, we tested GENEOnet robustness against non-rigid perturbations of the input data with the help of MD simulations. Our findings indicate that GENEOnet exhibits strong resilience to such perturbations, particularly when the perturbations are smaller, as in the case of 10 ps MD simulations. 

Regarding the robustness analysis, the reader could criticize that we considered only the initial segment of each MD simulation (in both cases of 100 ps and 10 ps time deltas) and that this could hide wider rearrangements of the protein structure that might happen later in the dynamics. We are aware of this fact and we plan to extend the analysis to the full MD simulations in order to prevent this eventuality and to reduce the variability in the mean overlaps estimates.

Considering all these aspects, the findings collectively illustrate that GENEOnet is a model that not only offers transparency but also maintains substantial reliability when subjected to geometric transformations and perturbations. Furthermore, in a broader sense, they highlight the value of GENEOs as a resource for creating innovative and trustworthy AI models with a focus on interpretability.

\section*{Acknowledgement(s)}

Scientific support is acknowledged by the Italian GNAMPA - INDAM group. Computational resources were partially provided by the INDACO core facility for HPC at Università degli Studi di Milano.

\section*{Data availability}
The two data sources used in the analyses are open and freely available at \href{http://pdbbind.org.cn/}{http://pdbbind.org.cn/} (for PDBbind) and \href{https://www.dsimb.inserm.fr/ATLAS/}{https://www.dsimb.inserm.fr/ATLAS/} (for ATLAS).

\section*{Disclosure statement}

Filippo Lunghini, Andrea Rosario Beccari, and Carmine Talarico are employees of Dompé Farmaceutici S.p.A.

\section*{Funding}

The authors acknowledge partial funding from Dompé Farmaceutici S.p.A. for developing this research.

\bibliographystyle{tfnlm}
\bibliography{references}

\section*{Appendices}

\setcounter{subsection}{0}
\renewcommand{\thesubsection}{\Alph{subsection}}
\renewcommand{\thetable}{\Alph{table}}

\subsection{Appendix on equivariance analysis}
\label{app:equivariance}

In this appendix we report Tables~\ref{tab:equi2},~\ref{tab:equi3} and~\ref{tab:equi4} containing equivariance proportion estimates for values of $\tau \in \{0.95, 0.75, 0.50\}$ which are shown in Figures~\ref{fig:equi1},~\ref{fig:equi2} and~\ref{fig:equi3} and also in the additional Table~\ref{tab:equi1} for the value of $\tau = 0.99$ that highlights a few mismatches due to numerical approximation also in the case of GENEOnet.

\begin{table}[h]
\begin{subtable}{\textwidth}
\centering
    \begin{tabular}{cccccc}
        \toprule
        Method $\mathcal{M}$ &  Pocket $j$ & Non-missing & $\hat{p}_j^{\mathcal{M};\tau}$ & $\text{SE}\bigl(\hat{p}_j^{\mathcal{M};\tau}\bigr)$ &    $CI\bigl(p_j^{\mathcal{M};\tau}\bigr)$\\
        \midrule
        \multirow{3}*{GENEOnet} & 1 & 2000 & 1.000000 & 0.000000 & [1.000000, 1.000000]\\
        & 2 & 1918 & 0.996350 & 0.001377 & [0.992803, 0.999898]\\
        & 3 & 1852 & 0.999460 & 0.000540 & [0.998069, 1.000851]\\
        \midrule
        \multirow{3}*{Deeppocket} & 1 & 1997 & 0.834251 & 0.008323 & [0.812812, 0.855691]\\
         & 2 & 1984 & 0.675907 & 0.010510 & [0.648834, 0.702980]\\
         & 3 & 1969 & 0.556120 & 0.011200 & [0.527271, 0.584968]\\
        \midrule
        \multirow{3}*{Fpocket} & 1 & 1998 & 0.642142 & 0.010727 & [0.614511, 0.669773]\\
        & 2 & 1987 & 0.367891 & 0.010821 & [0.340018, 0.395764]\\
        & 3 & 1971 & 0.263825 & 0.009929 & [0.238249, 0.289401]\\
        \midrule
        \multirow{3}*{P2Rank}  & 1 & 1866 & 0.007503 & 0.001998 & [0.002356, 0.012650]\\
         & 2 & 1441 & 0.022207 & 0.003883 & [0.012204, 0.032209]\\
         & 3 & 1078 & 0.022263 & 0.004496 & [0.010683, 0.033844]\\
        \bottomrule
    \end{tabular}
    \caption{\label{tab:equi1} $\tau = 0.99$}
\end{subtable}
\begin{subtable}{\textwidth}
\centering
    \begin{tabular}{cccccc}
        \toprule
        Method $\mathcal{M}$ &  Pocket $j$ & Non-missing & $\hat{p}_j^{\mathcal{M};\tau}$ & $\text{SE}\bigl(\hat{p}_j^{\mathcal{M};\tau}\bigr)$ &    $CI\bigl(p_j^{\mathcal{M};\tau}\bigr)$\\
        \midrule
        \multirow{3}*{GENEOnet} & 1 & 2000 & 1.000000 & 0.000000 & [1.000000, 1.000000]\\
         & 2 & 1918 & 1.000000 & 0.000000 & [1.000000, 1.000000]\\
         & 3 & 1852 & 1.000000 & 0.000000 & [1.000000, 1.000000]\\
        \midrule
        \multirow{3}*{Deeppocket} & 1 & 1997 & 0.838257 & 0.008242 & [0.817028, 0.859487]\\
        & 2 & 1984 & 0.679940 & 0.010476 & [0.652955,  0.706924]\\
        & 3 & 1969 & 0.559167 & 0.011192 & [0.530339,  0.587995]\\
        \midrule
        \multirow{3}*{Fpocket} & 1 & 1998 & 0.646146 & 0.010700 & [0.618584,  0.673708]\\
        & 2 & 1987 & 0.370408 & 0.010836 & [0.342495,  0.398320]\\
        & 3 & 1971 & 0.263825 & 0.009929 & [0.238249,  0.289401]\\
        \midrule
        \multirow{3}*{P2Rank} & 1 & 1866 & 0.008039 & 0.002068 & [0.002712,  0.013365]\\
        & 2 & 1441 & 0.022207 & 0.003883 & [0.012204,  0.032209]\\
        & 3 & 1078 & 0.022263 & 0.004496 & [0.010683,  0.033844]\\
        \bottomrule
    \end{tabular}
    \caption{\label{tab:equi2} $\tau = 0.95$}
\end{subtable}
\caption{Estimates of equivariance proportions.}
\end{table}

\begin{table}[ht]\ContinuedFloat
\begin{subtable}{\textwidth}
\centering
    \begin{tabular}{cccccc}
        \toprule
        Method $\mathcal{M}$ &  Pocket $j$ & Non-missing & $\hat{p}_j^{\mathcal{M};\tau}$ & $\text{SE}\bigl(\hat{p}_j^{\mathcal{M};\tau}\bigr)$ &    $CI\bigl(p_j^{\mathcal{M};\tau}\bigr)$\\
        \midrule
        \multirow{3}*{GENEOnet} & 1 & 2000 & 1.000000 & 0.000000 & [1.000000, 1.000000]\\
         & 2 & 1918 & 1.000000 & 0.000000 & [1.000000, 1.000000]\\
         & 3 & 1852 & 1.000000 & 0.000000 & [1.000000, 1.000000]\\
        \midrule
        \multirow{3}*{Deeppocket} & 1 & 1997 & 0.846770 & 0.008063 & [0.826002, 0.867538]\\
         & 2 & 1984 & 0.686492 & 0.010418 & [0.659657, 0.713327]\\
         & 3 & 1969 & 0.562722 & 0.011182 & [0.533920, 0.591525]\\
        \midrule
        \multirow{3}*{Fpocket} & 1 & 1998 & 0.649149 & 0.010679 & [0.621641, 0.676657]\\
        & 2 & 1987 & 0.373931 & 0.010857 & [0.345964, 0.401897]\\
        & 3 & 1971 & 0.265348 & 0.009948 & [0.239724, 0.290971]\\
        \midrule
        \multirow{3}*{P2Rank}  & 1 & 1866 & 0.379421 & 0.011236 & [0.350479, 0.408364]\\
         & 2 & 1441 & 0.186676 & 0.010268 & [0.160227, 0.213125]\\
         & 3 & 1078 & 0.128942 & 0.010212 & [0.102638, 0.155247]\\
        \bottomrule
    \end{tabular}
    \caption{\label{tab:equi3} $\tau = 0.75$}
\end{subtable}
\begin{subtable}{\textwidth}
\centering
    \begin{tabular}{cccccc}
        \toprule
        Method $\mathcal{M}$ &  Pocket $j$ & Non-missing & $\hat{p}_j^{\mathcal{M};\tau}$ & $\text{SE}\bigl(\hat{p}_j^{\mathcal{M};\tau}\bigr)$ &    $CI\bigl(p_j^{\mathcal{M};\tau}\bigr)$\\
        \midrule
        \multirow{3}*{GENEOnet} & 1 & 2000 & 1.000000 & 0.000000 & [1.000000, 1.000000]\\
         & 2 & 1918 & 1.000000 & 0.000000 & [1.000000, 1.000000]\\
         & 3 & 1852 & 1.000000 & 0.000000 & [1.000000, 1.000000]\\
        \midrule
        \multirow{3}*{Deeppocket} & 1 & 1997 & 0.854782 & 0.007886  & [0.834469, 0.875095]\\
         & 2 & 1984 & 0.692540 & 0.010362 & [0.665849, 0.719232]\\
         & 3 & 1969 & 0.566277 & 0.011171 & [0.537502, 0.595053]\\
        \midrule
        \multirow{3}*{Fpocket} & 1 & 1998 & 0.655155 & 0.010636  & [0.627758, 0.682553]\\
        & 2 & 1987 & 0.374937 & 0.010863 & [0.346956, 0.402918]\\
        & 3 & 1971 & 0.266362 & 0.009960 & [0.240708, 0.292017]\\
        \midrule
        \multirow{3}*{P2Rank}  & 1 & 1866 & 0.793676 & 0.009370 & [0.769540, 0.817813]\\
         & 2 & 1441 & 0.538515 & 0.013137 & [0.504676, 0.572354]\\
         & 3 & 1078 & 0.397032 & 0.014909 & [0.358628, 0.435435]\\
        \bottomrule
    \end{tabular}
    \caption{\label{tab:equi4} $\tau = 0.50$}
\end{subtable}
\caption{\label{tab:equi}Estimates of equivariance proportions.}
\end{table}

\subsection{Appendix on perturbation analysis}
\label{app:perturbation}

In this appendix, we report Table~\ref{tab:meanoverlap} containing sample mean overlaps for the various proteins analyzed in Section~\ref{subsec:robustness}. The Table reports estimates of mean overlaps both for the simulations of 100 ps ($\overline{O}_{100}$) and 10 ps ($\overline{O}_{10}$), their difference, and the $p$-value for testing $H_0\colon \mu_{100} = \mu_{10}$ against $H_1\colon \mu_{100} < \mu_{10}$ together with \verb|R| style significance codes. 

\begin{table}[ht]
\centering
\begin{tabular}{l|ccccl}
    \toprule      
    Protein  &  $\overline{O}_{100}$ & $\overline{O}_{10}$ & $\overline{O}_{100} - \overline{O}_{10}$ & $p$-value \\
    \midrule    
    16pk\_A & 0.877899 & 0.901982 & -0.024083 & 0.000240 & ***\\
    1d4t\_A & 0.791709 & 0.830869 & -0.039160 & 0.000955 & ***\\
    1fcz\_A & 0.898289 & 0.910334 & -0.012046 & 0.049147 & *\\
    1g6g\_A & 0.814539 & 0.827410 & -0.012872 & 0.154711 &  \\
    1ikt\_A & 0.840023 & 0.931655 & -0.091632 & 0.000000 & ***\\
    1jif\_A & 0.794048 & 0.818471 & -0.024423 & 0.050057 & .\\
    1n8u\_A & 0.880529 & 0.967725 & -0.087196 & 0.000000 & ***\\
    1nf8\_A & 0.868210 & 0.861657 & 0.006553 & 0.734692 &  \\
    1ntv\_A & 0.836641 & 0.861024 & -0.024383 & 0.005294 & **\\
    1ssq\_D & 0.808436 & 0.846528 & -0.038092 & 0.000138 & ***\\
    1ugp\_A & 0.839087 & 0.868306 & -0.029219 & 0.000913 & ***\\
    2e5y\_B & 0.871943 & 0.892541 & -0.020598 & 0.008978 & **\\
    2f5t\_X & 0.833802 & 0.845194 & -0.011392 & 0.106582 &  \\
    2hxm\_A & 0.825782 & 0.840013 & -0.014230 & 0.083565 & .\\
    2pmk\_A & 0.836137 & 0.870040 & -0.033903 & 0.000590 & ***\\
    2qmj\_A & 0.866217 & 0.879843 & -0.013627 & 0.041991 & *\\
    2vfk\_A & 0.893259 & 0.910764 & -0.017505 & 0.003343 & **\\
    2vmc\_A & 0.693255 & 0.788653 & -0.095397 & 0.000000 & ***\\
    2wr8\_A & 0.865839 & 0.887062 & -0.021224 & 0.005783 & **\\
    2x96\_A & 0.933093 & 0.941810 & -0.008716 & 0.006529 & **\\
    3d1g\_A & 0.827165 & 0.853228 & -0.026063 & 0.001443 & **\\
    3ebh\_A & 0.929064 & 0.933405 & -0.004341 & 0.139228 &  \\
    3g5k\_D & 0.934323 & 0.929183 & 0.005140 & 0.855327 &  \\
    3lbl\_A & 0.874922 & 0.905585 & -0.030663 & 0.000409 & ***\\
    3o0g\_D & 0.830042 & 0.834251 & -0.004209 & 0.358026 &  \\
    3p3g\_A & 0.910184 & 0.919091 & -0.008908 & 0.076540 & .\\
    3rt4\_A & 0.876687 & 0.893306 & -0.016619 & 0.009466 & **\\
    3v3l\_A & 0.844270 & 0.868580 & -0.024310 & 0.009285 & **\\
    4bt9\_B & 0.897038 & 0.926493 & -0.029454 & 0.000000 & ***\\
    4cjn\_A & 0.843593 & 0.894977 & -0.051384 & 0.000000 & ***\\
    4qxo\_A & 0.864800 & 0.881104 & -0.016304 & 0.014458 & *\\
    5ggl\_A & 0.797386 & 0.793500 & 0.003885 & 0.601223 &  \\
    5gof\_A & 0.851779 & 0.886590 & -0.034810 & 0.000000 & ***\\
    5l87\_A & 0.811304 & 0.841245 & -0.029941 & 0.009103 & **\\
    5ok6\_A & 0.824703 & 0.871197 & -0.046494 & 0.000000 & ***\\
    6bm5\_A & 0.836333 & 0.865143 & -0.028811 & 0.001632 & **\\
    6gfx\_C & 0.854570 & 0.880447 & -0.025876 & 0.001525 & **\\
    \bottomrule
    \end{tabular}
    \caption{\label{tab:meanoverlap}Mean overlap differences between 100 ps and 10 ps simulations. $p$-values are relative to testing $H_0\colon \mu_{100} = \mu_{10}$ against $H_1\colon \mu_{100} < \mu_{10}$.}
\end{table}

\end{document}